\documentclass[letterpaper, 10 pt, journal, twoside]{IEEEtran}

\IEEEoverridecommandlockouts                              

\usepackage{graphicx} 
\usepackage{amsmath} 
\usepackage{amssymb}  
\usepackage{makecell}
\usepackage{physics}
\usepackage[ruled, vlined, linesnumbered, nofillcomment]{algorithm2e}
\usepackage{array}
\usepackage{tabularx}
\usepackage{adjustbox}
\usepackage{makecell}
\usepackage{multicol}
\usepackage{bbm}
\usepackage{multirow}
\usepackage[bookmarks=true,hidelinks]{hyperref} 
\usepackage[caption=false]{subfig}
\usepackage{booktabs}
\usepackage{comment}
\usepackage{bm}
\usepackage{threeparttable}
\usepackage{cleveref}
\usepackage[table,xcdraw,dvipsnames]{xcolor}
\usepackage{hhline}
\usepackage{arydshln} 
\usepackage{cancel}
\usepackage{graphbox} 
\usepackage{tikz}
\usepackage{listofitems} 
\usetikzlibrary{calc}

\DeclareMathAlphabet{\mathpzc}{OT1}{pzc}{m}{it}

\DeclareMathOperator*{\argmin}{arg\,min}

\newcommand{\eg}{{\em e.g.}}
\newcommand{\ie}{{\em i.e.}}

\newcommand{\reviewdiff}[1]{#1}
\newcommand{\maybevspace}[1]{}
\newcommand{\fixTableTopMargin}{\vspace{+3pt}}

\usepackage{cellspace}
\usepackage{scalerel}
\usetikzlibrary{svg.path}
\definecolor{orcidlogocol}{HTML}{A6CE39}
\tikzset{
    orcidlogo/.pic={
        \fill[orcidlogocol] svg{M256,128c0,70.7-57.3,128-128,128C57.3,256,0,198.7,0,128C0,57.3,57.3,0,128,0C198.7,0,256,57.3,256,128z};
        \fill[white] svg{M86.3,186.2H70.9V79.1h15.4v48.4V186.2z}
        svg{M108.9,79.1h41.6c39.6,0,57,28.3,57,53.6c0,27.5-21.5,53.6-56.8,53.6h-41.8V79.1z M124.3,172.4h24.5c34.9,0,42.9-26.5,42.9-39.7c0-21.5-13.7-39.7-43.7-39.7h-23.7V172.4z}
        svg{M88.7,56.8c0,5.5-4.5,10.1-10.1,10.1c-5.6,0-10.1-4.6-10.1-10.1c0-5.6,4.5-10.1,10.1-10.1C84.2,46.7,88.7,51.3,88.7,56.8z};
    }
}

\DeclareFontFamily{OT1}{pzc}{}
\DeclareFontShape{OT1}{pzc}{m}{it}{<-> s * [1.05] pzcmi7t}{}
\DeclareMathAlphabet{\mathpzc}{OT1}{pzc}{m}{it}

\newcommand\orcidicon[1]{\href{https://orcid.org/#1}{\mbox{\scalerel*{
                \begin{tikzpicture}[yscale=-1,transform shape]
                \pic{orcidlogo};
                \end{tikzpicture}
            }{|}}}}

\usepackage{background}
\usepackage{stackengine}
\setstackEOL{\\}
\setstackgap{L}{\normalbaselineskip}

\SetBgScale{1}
\SetBgContents{\normalsize\stackunder[37cm]{\Longstack{%
This article has been accepted for publication in IEEE Robotics and Automation Letters (RA-L). This is the author's version which has not been fully edited and\\content may change prior to final publication. Citation information: DOI 10.1109/LRA.2025.3534683 }}{%
		\rotatebox{0}{\textcopyright \hspace{1ex}2025 IEEE. Personal use is permitted, but republication/redistribution requires IEEE permission.
			See \url{https://www.ieee.org/publications/rights/index.html} for more information.}}}
\SetBgColor{black}
\SetBgAngle{0}
\SetBgOpacity{0.7}
\SetBgScale{0.7}

\colorlet{colorFst}{Green!25}       
\colorlet{colorSnd}{SpringGreen!35}  
\colorlet{colorTrd}{Yellow!15}      
\colorlet{colorLow}{darkgray!30}    
\newcommand{\markfst}[1]{\cellcolor{colorFst}\bf{#1}}   
\newcommand{\marksnd}[1]{\cellcolor{colorSnd}{#1}}      
\newcommand{\marktrd}[1]{{#1}}      

\title{
GEVO: Memory-Efficient Monocular Visual Odometry Using Gaussians
}

\author{Dasong Gao*$^{\textsuperscript{\orcidicon{0000-0002-1391-0869}}}$, Peter Zhi Xuan Li*$^{\textsuperscript{\orcidicon{0000-0002-5260-4995}}}$, Vivienne Sze$^{\textsuperscript{\orcidicon{0000-0003-4841-3990}}}$, Sertac Karaman$^{\textsuperscript{\orcidicon{0000-0002-2225-7275}}}$
\thanks{Manuscript received: September, 13, 2024; Revised November, 27, 2024; Accepted January, 21, 2025.
This paper was recommended for publication by Editor J. Civera upon evaluation of the Associate Editor and Reviewers' comments.
This work was supported by MIT-MathWorks Fellowship, Amazon and NSF CPS 1837212.} 
\thanks{The first two authors contributed equally to this work. Authors are with the Massachusetts Institute of Technology, Cambridge, MA  02139,  USA.  Emails: {\tt\{dasongg, peterli, sze, sertac\}@mit.edu}.}	
\thanks{Digital Object Identifier (DOI): 10.1109/LRA.2025.3534683.}
}

\begin{document}

\maketitle


\begin{abstract}
Constructing a high-fidelity representation of the 3D scene using a monocular camera can enable a wide range of applications on low-energy devices, such as micro-robots, smartphones, and AR/VR headsets.
On these devices, memory is often limited in capacity and its access often dominates the consumption of compute energy.
Although Gaussian Splatting (GS) allows for high-fidelity reconstruction of 3D scenes, current GS-based SLAM is not memory efficient as a large number of past images is stored to retrain Gaussians for reducing catastrophic forgetting.
These images often require two-orders-of-magnitude higher memory than the map itself and thus dominate the total memory usage.
In this work, we present GEVO, a GS-based monocular SLAM framework that achieves comparable fidelity as prior methods by rendering (instead of storing) them from the existing map.
%
Novel Gaussian initialization and optimization techniques are proposed to remove artifacts from the map and delay the degradation of the rendered images over time.
%
%
%
%
%
Across a variety of environments, GEVO achieves comparable map fidelity while reducing the memory overhead to around 58~MBs, which is up to $94\times$ lower than prior works.
\end{abstract}


\section{Introduction}
\IEEEPARstart{E}nergy-constrained devices, such as AR/VR headsets and micro-robots, enable a wide range of applications that involve long-term and safe interactions with 3D environments through their high fidelity 3D representation.
%
%
\reviewdiff{For instance, constructing 3D representation in real-time allows i) AR/VR headsets to warn users when they are too close to the real-world obstacles obscured by virtual ones, and ii) micro-robots to perform collision checking and path planning during autonomous exploration.}
%
%
Due to limited battery capacity, these devices rely on low-power passive sensors to perceive the environment.
Thus, it is crucial for online simultaneous localization and mapping (SLAM) to produce a \emph{high fidelity} 3D map with a \emph{monocular} RGB camera.

On energy-constrained devices, memory is often limited in capacity and its access could dominate the total compute energy.
For instance, the energy for accessing data in an 8~KB cache is $10\times$ more than performing a 32-bit floating point multiplication~\cite{markhoro}.
Furthermore, the energy for accessing data in off-chip DRAM (GBs of capacity) is orders of magnitude higher than that for on-chip cache (KBs to MBs of capacity)~\cite{markhoro}.
Thus, algorithms should be \emph{memory-efficient} with low memory usage so that variables can be effectively cached on chip to reduce energy.
%
%

\begin{figure}[t]
    \centering
    \captionsetup[subfloat]{justification=centering}
    \tikzstyle{psnr_textbox}=[anchor=south east,rectangle,preaction={fill=white,opacity=0.6},font={\fontsize{6pt}{12}\bfseries},inner sep=2]
    \tikzstyle{dotted_box}=[ultra thick, densely dotted]
    \subfloat[][Reconstruction before catastrophic forgetting]{%
        \begin{tikzpicture}
            \node (fig) [anchor=south west,inner sep=0] at (0,0) {\includegraphics[width=0.45\columnwidth]{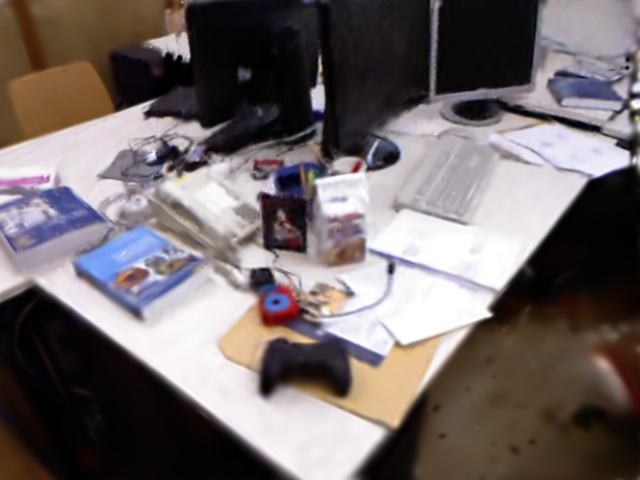}};
            \node[psnr_textbox] at ($(fig.south east)+(-0.02, 0.02)$) {PSNR: 27.5~dB};
        \end{tikzpicture}
        \label{fig:forgetting-visualization-after-current}
    }
    \subfloat[][MonoGS~\cite{matsuki2024gaussian} no past images\\ 8 images stored (7 MB)]{%
        \begin{tikzpicture}
            \node (fig) [anchor=south west,inner sep=0] at (0,0) {\includegraphics[width=0.45\columnwidth]{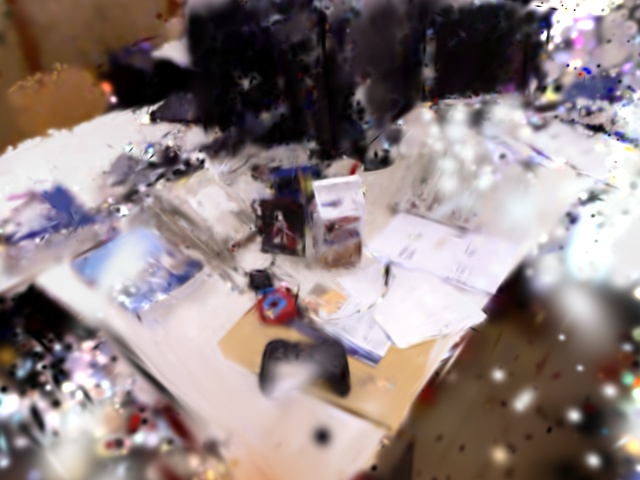}};
            \draw[dotted_box, green] (0.05, 1.8) rectangle (1.1, 2.7);
            \draw[dotted_box, green] (0.05, 0.05) rectangle (0.9, 0.9);
            \draw[dotted_box, magenta] (2.5, 1.6) rectangle (3.3, 2.4);
            \draw[dotted_box, magenta] (1.5, 0.05) rectangle (2.1, 0.9);
            \node[psnr_textbox] at ($(fig.south east)+(-0.02, 0.02)$) {PSNR: 11.8~dB};
        \end{tikzpicture}
        \label{fig:forgetting-visualization-last}
    }
    \\ [+1ex]
    \subfloat[][{GEVO (This work)} \\ 8 images stored (7 MB)]{%
        \begin{tikzpicture}
            \node (fig) [anchor=south west,inner sep=0] at (0,0) {\includegraphics[width=0.45\columnwidth]{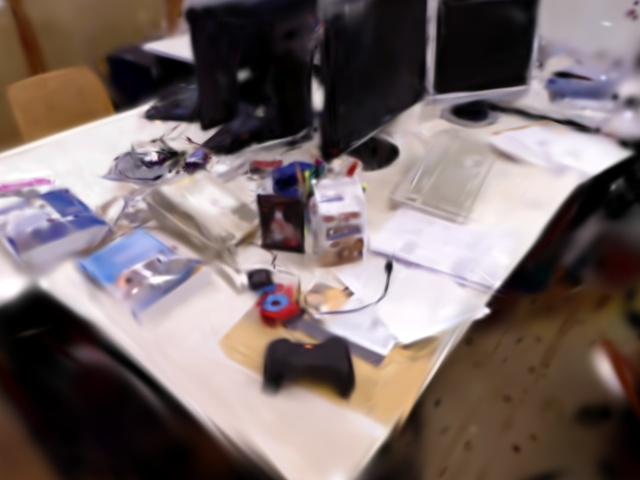}};
            \node[psnr_textbox] at ($(fig.south east)+(-0.02, 0.02)$) {PSNR: 18.93~dB};
        \end{tikzpicture}
        \label{fig:forgetting-visualization-ours}
    }
    \subfloat[][MonoGS~\cite{matsuki2024gaussian} \\ 114 images stored (100 MB)]{%
        \begin{tikzpicture}
            \node (fig) [anchor=south west,inner sep=0] at (0,0) {\includegraphics[width=0.45\columnwidth]{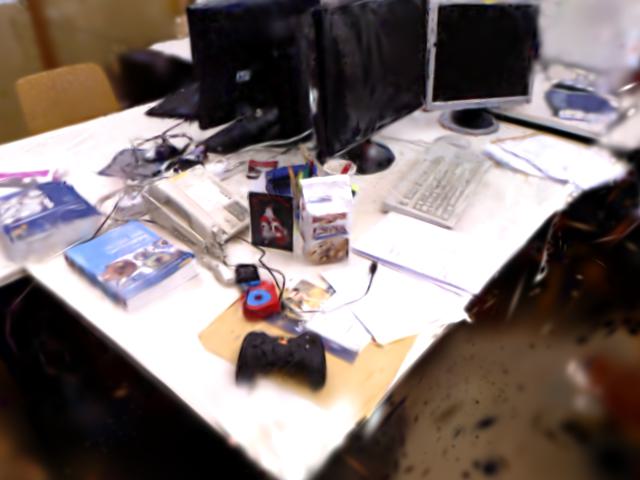}};
            \node[psnr_textbox] at ($(fig.south east)+(-0.02, 0.02)$) {PSNR: 19.68~dB};
        \end{tikzpicture}
        \label{fig:forgetting-visualization-photo-slam}
    }
    \caption{During online GS-based SLAM, the map \reviewdiff{(consisting of 3D Gaussians) is built by rendering and optimizing at each viewpoint} using a sliding window buffer of images. 
    a)~The region visible during the current sliding window achieves high fidelity after initial optimization.
    b)~However, without storing and retraining the map on a large number of past images, the fidelity of the same region degrades over time due to forgetting (artifacts in rectangles).
    c)~While alleviating forgetting, our GEVO avoids storing past images to reduce the memory overhead.
    d)~To achieve similar map fidelity, MonoGS~\cite{matsuki2023gaussian} stores all past keyframes and incurs a memory overhead at least 50$\times$ higher than the size of the map.
    }
    \label{fig:forgetting-visualization}
    \maybevspace{-20pt}
\end{figure}

Constructing a high-fidelity map requires RGB images to guide its optimization process.
To achieve real-time operation, many SLAM frameworks~\cite{campos2021orb, forster2016manifold, davison2007monoslam, Mur-Artal-RSS-15} optimize the pose and map using a small sliding window of images.
However, for dense SLAM, the map tends to catastrophically forget and degrade over time after the sliding window has passed (see \Cref{fig:forgetting-visualization-after-current} vs. \ref{fig:forgetting-visualization-last}).
%
%
%
%
To alleviate forgetting, 
prior frameworks \emph{additionally} store a large number of past images outside the current sliding window to repeatedly retrain the map.
\reviewdiff{These frameworks use either neural networks~\cite{rosinol2023nerf, matsuki2023imode, zhu2024nicer} or learnable 3D Gaussians~\cite{matsuki2023gaussian, huang2023photo} to map both occupancy and color of the environment.}
Unfortunately, the overhead memory used to store these images dominates the total memory and is orders of magnitude higher than both the current sliding window and the map itself.

\begin{figure}[t]
    \centering
    \subfloat[][\textbf{RO:} The current view inconsistently inserts or moves a {\color{orange} new Gaussian} into the {\color{CadetBlue} obstacle-free region} of previous views to \emph{occlude} the {\color{red} existing Gaussians}.]{%
        \includegraphics[width=0.47\columnwidth]{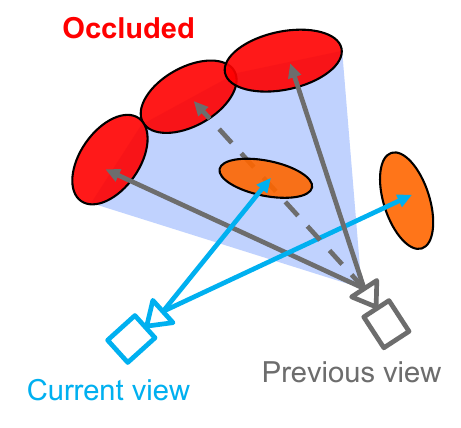}
        \label{fig:ro}
    }
    \hfill
    \subfloat[][\textbf{IRO}: Sensor rays of the current view pass through {\color{orange} new Gaussians} to cause \emph{overfitting} of an {\color{red} existing Gaussian} created from previous view.]{%
        \includegraphics[width=0.43\columnwidth]{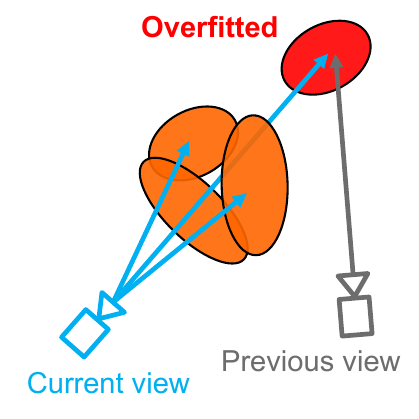}
        \label{fig:iro}
    }
	\caption{Two scenarios that cause catastrophic forgetting in Gaussian Splatting: b) retrospective occlusion (RO) and a) incomplete ray obscuration (IRO). RO causes the new Gaussians to occlude ones in the past view (red rectangles in \Cref{fig:forgetting-visualization-last}). IRO causes the existing Gaussians to overfit to the current view (green rectangles in \Cref{fig:forgetting-visualization-last}).
		}
	\label{fig:forgetting} 
    \maybevspace{-16pt}
\end{figure}

Our contribution is a memory-efficient SLAM framework using Gaussians, called GEVO, that significantly reduces memory overhead by \emph{rendering} past images from the existing map instead of storing them in memory.
However, the fidelity of these images is lower than the original and can slowly degrade overtime due to the artifacts in the map caused by forgetting.
Thus, using these images to guide Gaussian optimization via splatting (\ie, GS~\cite{kerbl20233d}) alone is insufficient for constructing a high fidelity map.
To complement GS, GEVO contains the following procedures to further reduce incorrect occlusion and overfitting due to catastrophic forgetting:
%
\begin{enumerate}
    \item \textbf{Occupancy-Preserving Initialization}: To reduce incorrect occlusions, Gaussians that lie within the obstacle-free regions (\ie, orange Gaussian in the blue region of \Cref{fig:ro}) are pruned. Thus, in addition to representing obstacles, Gaussians representing free regions are initialized to identify incorrect occlusions.
    %
	\item \textbf{Consistency-Aware Optimization:} To reduce overfitting of the map to the current window, we only optimize a small subset of Gaussians that are both inconsistent and sufficiently visible to the camera (see orange Gaussians in \Cref{fig:iro}). To ensure rendered images maintain high fidelity, we locally optimize noisy Gaussians created from the current sliding window before merging them to the map for global optimization.
\end{enumerate}

Across a variety of environments, GEVO achieves comparable map fidelity (see \Cref{fig:forgetting-visualization-ours}) and reduces the memory overhead to around 58MBs, which is up to $94\times$ lower than prior works. Thus, GEVO makes a significant stride towards the deployment of GS-based SLAM on low-energy devices.
\section{Related Work}
%
Monocular SLAM frameworks can be classified based on the type of scene representation (\eg, points, planes, neural networks, Gaussians).
Constructing these representations online exhibits different trade-offs in efficiency and fidelity.

\textbf{Traditional Monocular SLAM:} Traditional SLAM frameworks can be classified as \emph{indirect} or \emph{direct}, and excel at real-time localization by tracking and optimizing over a set of points representing the 3D scene.
For \emph{indirect} frameworks~\cite{campos2021orb, forster2016manifold, davison2007monoslam, Mur-Artal-RSS-15}, the set of points are selected using feature extractors~\cite{lowe2004sift, rublee2011orb} that seek to uniquely identify certain characteristics of the environment such as corners.
Although these frameworks are often memory-efficient and real-time, the amount of unique features is very sparse which produces a map with very low coverage of the environment.
%
In contrast, \emph{direct} frameworks~\cite{engel2014lsd, dtam, ptam, dso} seek to track a denser set of points with high photometric gradients at the expense of larger memory and computational overhead.
Although the coverage of the environment increases, reconstruction is not photo-realistic and contains significant noise in regions with less texture~\cite{Mur-Artal-RSS-15}.

To reduce sparsity, other traditional frameworks explore more descriptive geometric primitives, such as planes~\cite{yang2019monocular}, quadrics~\cite{nicholson2018quadricslam} and meshes~\cite{rosinol2020kimera}.
However, these frameworks coarsely track the locations of objects that conform to their respective primitives and thus struggle with modeling remaining objects in the scene that do not conform.

\textbf{Neural Monocular SLAM:} To provide a photo-realistic reconstruction of the environment, neural-based frameworks, such as GO-SLAM~\cite{zhang2023go}, NICER-SLAM~\cite{zhu2024nicer}, and iMODE~\cite{matsuki2023imode} represent the environment using a Neural Radiance Field (NeRF).
Due to the volumetric rendering required for training NeRFs, the training process is computationally intensive.
%
Thus, most of these frameworks propose techniques that accelerate training, some of which
include i) using a hybrid scene representation with the voxel grids (in NICER-SLAM) or hash table (in GO-SLAM), and ii) training on a carefully selected subset of input images (in most prior works including iMODE).

Even though throughput was enhanced by these techniques, almost all Neural SLAM frameworks suffer from \emph{catastrophic forgetting}, which is reduced by periodic re-training on images acquired throughout the entire experiment.
%
Thus, these images need to be stored as overhead in memory, which quickly grows with the duration of the experiment to dominate the total memory usage.
%

\textbf{Gaussian Monocular SLAM:} To improve throughput and achieve photo-realistic rendering, recent frameworks, such as MonoGS~\cite{matsuki2024gaussian}, Photo-SLAM~\cite{huang2024photo}, and SplatSLAM~\cite{keetha2024splatam}, use Gaussian Splatting (GS) to train learnable Gaussians for 3D representation.
There frameworks propose different localization techniques to complement GS.
For instance, both MonoGS and SplatSLAM localize the camera against the global map via minimizing a photometric cost function, while Photo-SLAM utilizes ORB-SLAM~\cite{campos2021orb}.
%

Similar to Neural SLAM, current Gaussian SLAM frameworks also suffer from \emph{catastrophic forgetting} and thus require the storage of a large number of images to periodically retrain all Gaussians.
%
%
%
In this work, we propose memory-efficient techniques that reduce catastrophic forgetting in GS-based frameworks by \emph{rendering} most images from the map instead of storing them in memory.
From \Cref{fig:forgetting-visualization}, our framework, named GEVO, can also achieve comparable rendering accuracy while requiring significantly less memory overhead compared with prior frameworks.
%

%
%

\begin{figure*}[t]
	\centering
	\includegraphics[width=0.8\textwidth]{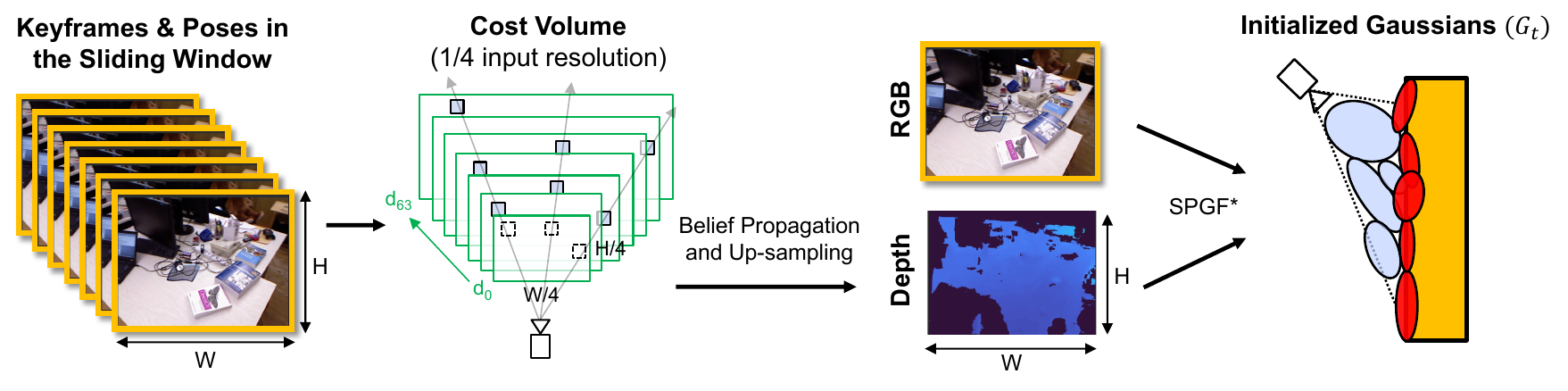}
	\caption{\textbf{Occupancy-Preserving Initialization.} Given a set of recently acquired keyframes and poses in a sliding window buffer, the depth image for the most recent keyframe is computed using belief propagation on a photometric cost volume at a quarter of the image's resolution.
	Then, the depth and RGB image are used to initialize a set of Gaussians ($\mathcal{G}_t$) for representing obstacles (red) and free region (blue) using the SPGF* algorithm~\cite{gmmap}.
	Gaussians representing free regions are fused across multiple keyframes to identify instances of retrospective occlusion (RO) during consistency-aware optimization.}
	\label{fig:gaussian_initialization}
    \maybevspace{-12pt}
\end{figure*}

\begin{figure*}[t]
	\centering
	\includegraphics[width=\textwidth]{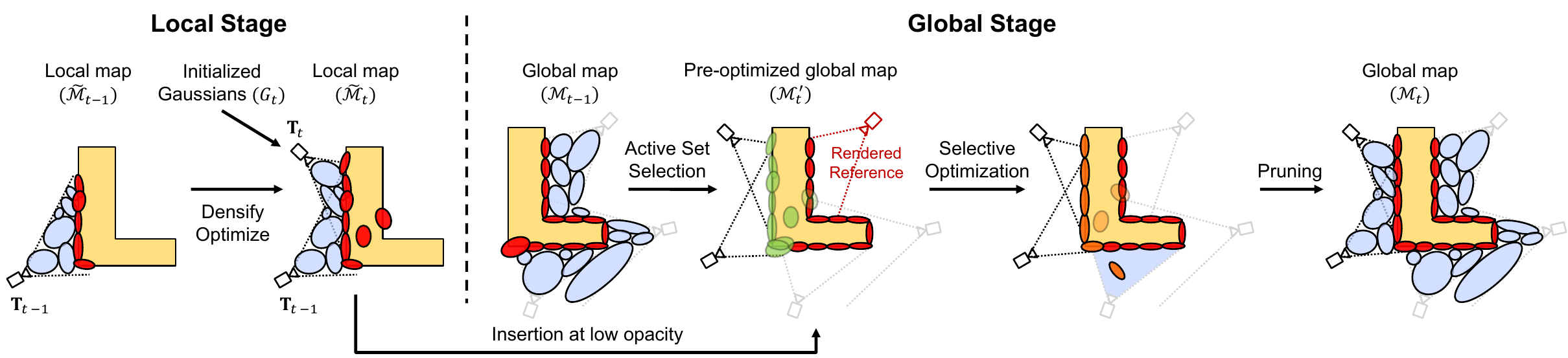}
        \caption{\textbf{Consistency-Aware Optimization.} Given newly initialized Gaussians ($\mathcal{G}_t$), we perform a GS-based optimization in two stages: i) \emph{Local stage} performs GS to optimize a local map $\widetilde{\mathcal{M}}_{t}$ that represents all geometries visible from the sliding window, and ii) \emph{Global stage} selectively optimizes a small active set $\mathcal{A}_t$ of Gaussians (green) consisting of the local map $\widetilde{\mathcal{M}}_{t}$ and existing Gaussians $\mathcal{E}_t$ with high rendering error. 
       	This active set selection tends to exclude Gaussians obscured from camera views used for training and thus reduces IRO.
        Since Gaussians from the local map are sufficiently accurate, images from randomly selected past views are \emph{rendered} from the global map to guide the global optimization stage. Finally, Gaussians that causes RO are pruned with the help of the obstacle-free regions created during initialization (blue). Note that Gaussians representing free regions are omitted in the global stage except for the pruning step for ease of visualization.}
	\label{fig:map_fusion}
    \maybevspace{-16pt}
\end{figure*}


\section{Proposed Methods}
In this section, we present GEVO, a memory-efficient GS-based monocular SLAM framework, that reduces \emph{catastrophic forgetting} by relying on images \emph{rendered} from the map to guide the GS optimization process.
Recall that online SLAMs operate on a sliding window of images for localization and mapping.
Catastrophic forgetting can occur when Gaussians created from the current window \emph{occlude} the ones from the past windows.
As illustrated in \Cref{fig:ro}, the inconsistency is caused by retrospective occlusion (RO) where Gaussians from the current view (orange) lie within the obstacle-free region (blue) of the prior view.
%

Catastrophic forgetting can also occur when a previously observed region from a past sliding window \emph{overfit} to images from the current sliding window.
As illustrated in \Cref{fig:iro}, overfitting is caused by incomplete ray obscuration (IRO) where Gaussians that are associated with the current view (red) do not completely obscure the sensor rays (blue).
Thus, Gaussians created from prior views (orange) are still partially visible such that their parameters update to match the appearances in the current view.
%
%
In prior works, both RO and IRO are reduced by storing images from all sliding windows to retrain the Gaussians.
These images typically dominate the total memory usage which grows over time.

Using \emph{rendered} images from the existing map is not sufficient for reducing catastrophic forgetting during optimization.
In particular, the fidelity of these images slowly degrades over time due to artifacts in the map caused by forgetting from RO and IRO.
%
%
To reduce RO, our framework consists of an accurate Gaussian initialization procedure (\Cref{subsec:initialization}) that compactly encodes obstacle-free regions. These regions are used to identify instances where new Gaussians occlude existing ones, which are pruned at the end of optimization.
To reduce IRO, we propose a two-stage optimization procedure (\Cref{subsec:optimization}) to update a small subset of Gaussians that are both inconsistent and sufficiently visible to the current sliding window so that the remaining Gaussians do not overfit to the current window.
%

\subsection{Occupancy-Preserving Initialization}~\label{subsec:initialization}
In this section, we present an efficient procedure that initializes Gaussian parameters representing obstacles and obstacle-free regions in the current sliding window.
%
%
%
The free regions are used to prune Gaussians that causes RO.

To achieve efficiency and good generalization across a variety of environments, our procedure is adapted from an efficient implementation of multi-view stereo (MVS)~\cite{wang2018quadtree} and does not rely on time-consuming COLMAP~\cite{schoenberger2016sfm, schoenberger2016mvs} or less accurate random sampling~\cite{matsuki2024gaussian}.
\Cref{fig:gaussian_initialization} summarizes our procedure.
Given a sequence of RGB keyframes in a sliding window, we construct a cost volume that captures the photometric consistency for each pixel in the most recent image at different depth hypotheses.
From the cost volume, belief propagation~\cite{felzenszwalb2006efficient} is performed to extract a depth image associated with the most recent image.
Finally, Gaussian parameters for both obstacles (red) and free regions (blue) are computed by the memory-efficient SPGF* algorithm~\cite{gmmap}.

Our procedure can be integrated with many localization and keyframe selection strategies, such as map-centric direct methods in MonoGS~\cite{matsuki2024gaussian} and feature-based methods in ORB-SLAM~\cite{campos2021orb}.
Details about our procedure are described below.

\subsubsection{Cost Volume Generation}~\label{subsec:cv_gen}
Given a sequence of $N =$ 8 or 10 keyframes $(I_0, \dots, I_{N-1})$ from the sliding window buffer (with $I_{N-1}$ being the most recent), the value of the photometric cost $V(\mathbf{u}, d)$ for pixel $\mathbf{u}$ in $I_{N-1}$ when depth is hypothesized to be $d$ is defined as
\begin{equation}\label{eqn:photometric_cost}
	V(\mathbf{u}, d) = \frac{1}{N-1}\sum_{i = 0}^{N-2} \lvert I_{N-1}(\mathbf{u}) -  I_i\left(\pi(\mathbf{u}, d, \mathbf{T}^i_{N-1})\right) \rvert,
\end{equation}
where $I(\cdot)$ is the intensity of a specific pixel in image $I$, $\mathbf{T}^i_{N-1} \in \mathbb{SE}(3)$ is the transformation matrix from image $I_{N-1}$ to $I_{i}$, $\pi(\cdot)$ warps the coordinate $\mathbf{u}$ from image $I_{N-1}$ to $I_{i}$ given the depth hypotheses $d$.

In our experiments, we choose 64 depth hypothesis (\ie, $\{d_0, \dots, d_{63}\}$ that are equally spaced from $0.25 - 25$ m. 
To reduce memory overhead, we downsample the image by $4\times$ in each dimension before creating the cost volume in order to exploit the spatial redundancy in the image. 
Assuming that each image has height $H = 480$ and width $W=640$, the resulting cost volume $V$ only requires 4.7 MBs. 
%
%
\reviewdiff{Since we use \textit{full-resolution RGB images} for subsequent optimizations, potential loss in spatial details will be recovered.}

\subsubsection{Gaussian Generation}~\label{subsec:gau_gen}
%
%
To determine the most likely depth hypothesis for each pixel under the assumption that each obstacle has smooth surfaces, belief propagation (BP) from~\cite{felzenszwalb2006efficient} is used to extract a depth image for the most recent keyframe in the sliding window buffer.
Under the assumption that neighboring pixels with the same color in the keyframe are likely to describe the same surface, we use an efficient algorithm proposed in~\cite{min2014fast} to upsample the depth image from BP to the full resolution of the keyframe. 

Given the most recent keyframe and its depth image, we enhanced a memory-efficient algorithm, called SPGF*~\cite{gmmap}, to generate a set of Gaussians representing obstacles (red) and free (blue) regions (see \Cref{fig:gaussian_initialization}).
Unlike prior approaches~\cite{eckart2016accelerated, o2018variable, dhawale2020efficient, goel2023probabilistic} that process the depth image in multiple passes, SPGF* exploits the connectivity encoded in the depth image to efficiently generate the Gaussians in a single pass with comparable accuracy.
Since SPGF* was mainly designed for accurate depth reconstruction, each Gaussian that represents an obstacle could enclose a surface containing multiple colors.
To enhance the fidelity of color representation, we modified SPGF* to ensure that each Gaussian can only represent a surface with a similar color.
%

\subsection{Consistency-Aware Optimization}~\label{subsec:optimization}
After Gaussians are initialized in \Cref{subsec:initialization}, they are fused into the global map as illustrated in \Cref{fig:map_fusion}.
%
%
Two challenges arise when integrating the new Gaussians into the existing map using the current sliding window:
i) Since these Gaussians are potentially noisy, inserting them into the map likely causes RO;
ii) During optimization, both IRO and RO tend to occur due to the lack of constraints from past views.
In prior works~\cite{matsuki2024gaussian, huang2024photo, zhang2023go}, both are resolved by training Gaussians on keyframes sampled from past sliding windows.
In our work, we rely on past keyframes \emph{rendered} from the existing map to reduce memory overhead.
%
%

However, the fidelity of these rendered images degrades over time which leads to significant degradation of the map itself.
To maintain fidelity, we employ a two-stage optimization that first enhances the initialized Gaussians in the local stage before optimizing them with other existing Gaussians in the global stage.
Since existing Gaussians are not perturbed during the local stage, they are used to render past keyframes with high fidelity.
To reduce IRO, we optimize a small subset of Gaussians that are both inconsistent and visible from the current sliding window.
In addition, the free regions encoded by Gaussians from past keyframes allow us to further reduce RO via occupancy-based pruning.
%

\subsubsection{Local Stage}~\label{subsubsec:local-mapping}
The initialized Gaussians $\mathcal{G}_t$ are appended with Gaussians $\widetilde{\mathcal{M}}_{t-1}$ that are also visible from the current sliding window to form an updated local map $\widetilde{\mathcal{M}}_{t}$. 
To reduce the noise of the update map caused by recently initialized Gaussians, 
%
the map is optimized using images from only the current sliding window via the objective~\cite{matsuki2024gaussian},
%
\begin{equation} \label{eqn:local-mapping}
    \argmin_{\mathbf{T}_k, \widetilde{\mathcal{M}}_{t}} E_{pho} + E_{iso},
\end{equation}
%
where $E_{pho}$ is the photometric loss between rendered and ground truth images, $E_{iso}$ is the isotropic loss, which prevents the formation of elongated or thin Gaussians, and $\mathbf{T}_k \in \mathbb{SE}(3)$ is its estimated pose.

%

\begin{figure*}[t]
    \tikzstyle{dotted_box}=[ultra thick, densely dotted]
    \newcommand{\drawdottedbox}[2]{
        \setsepchar{;/,}
        \readlist*\boxlist{#1}
        \begin{tikzpicture}
            \node (fig) [anchor=south west,inner sep=0] at (0,0) {#2};
            \foreachitem\x\in\boxlist[]{
                \draw[dotted_box, {\boxlist[\xcnt,5]}] (\boxlist[\xcnt,1], \boxlist[\xcnt,2]) rectangle (\boxlist[\xcnt,3], \boxlist[\xcnt,4]);
            }
        \end{tikzpicture}
    }
    \newcommand{\reprects}{0.05, 0.3, 1.5, 1.7, green}
    \newcommand{\reprectsl}{2.8, 1.3, 3.5, 2.0, green; 0.05, 0.3, 1.5, 1.7, green}
    \newcommand{\tumrects}{0.5, 0.2, 1.1, 0.8, magenta; 0.6, 2.2, 3.2, 2.65, green}
    
    \centering
    \fixTableTopMargin
    \fixTableTopMargin
    \resizebox{\linewidth}{!}{
        \setlength{\tabcolsep}{0pt}
        \begin{tabular}{cccccc}
            %
            %
            \drawdottedbox{\reprects}{\includegraphics[width=0.2\linewidth,trim={0px 0px 293px 0px},clip]{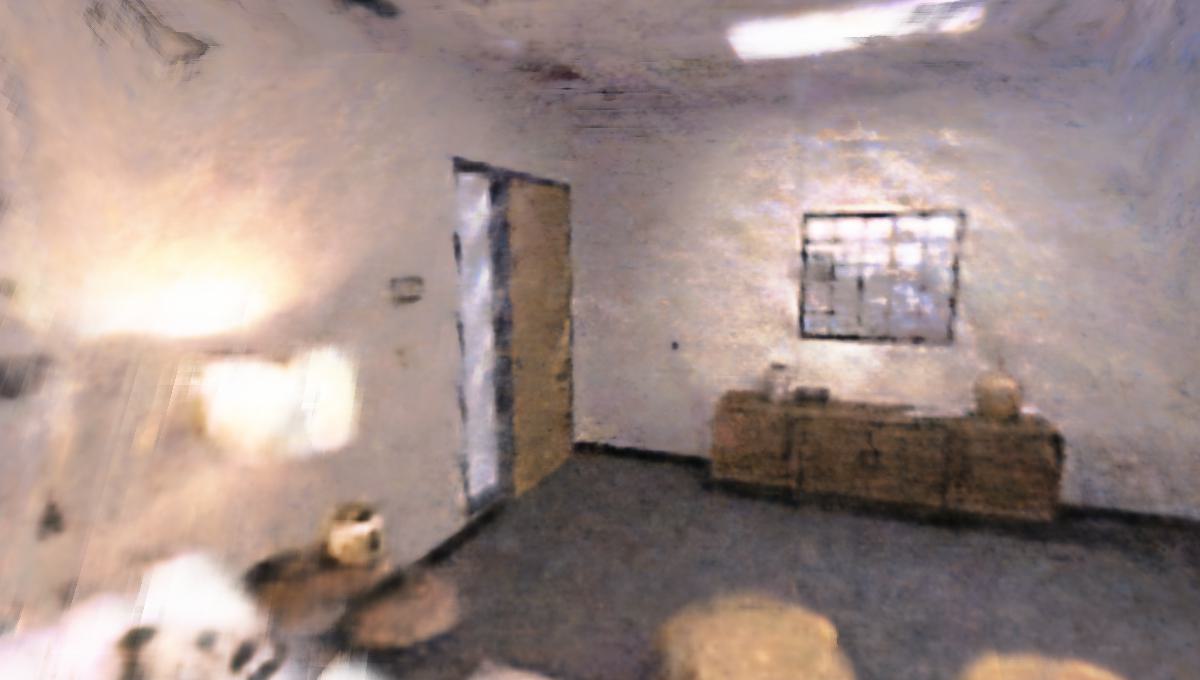}} &
            \drawdottedbox{\reprects}{\includegraphics[width=0.2\linewidth,trim={0px 0px 293px 0px},clip]{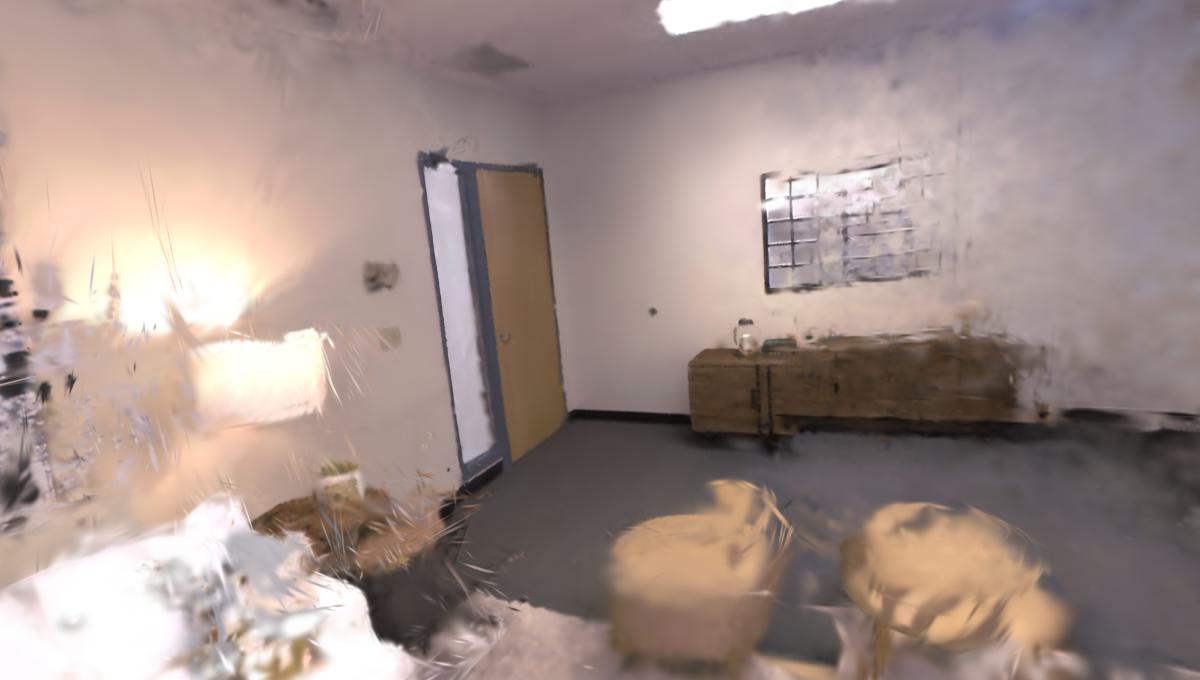}} &
            \drawdottedbox{\reprects}{\includegraphics[width=0.2\linewidth,trim={0px 0px 293px 0px},clip]{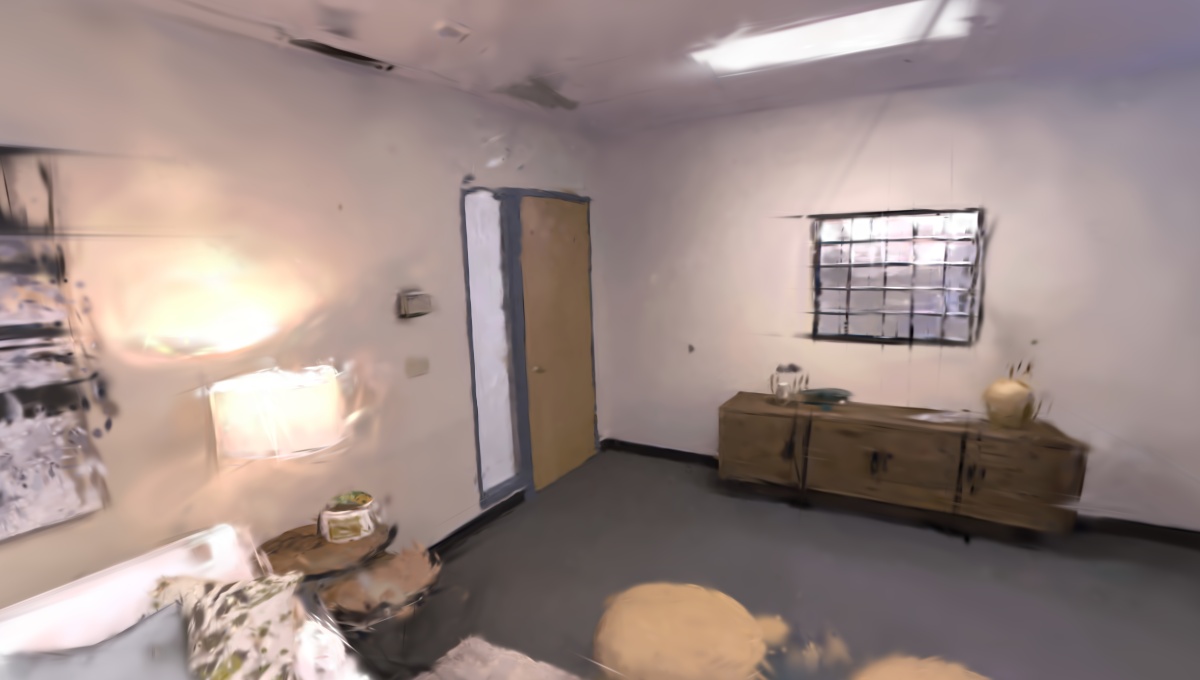}} &
            \drawdottedbox{\reprects}{\includegraphics[width=0.2\linewidth,trim={0px 0px 293px 0px},clip]{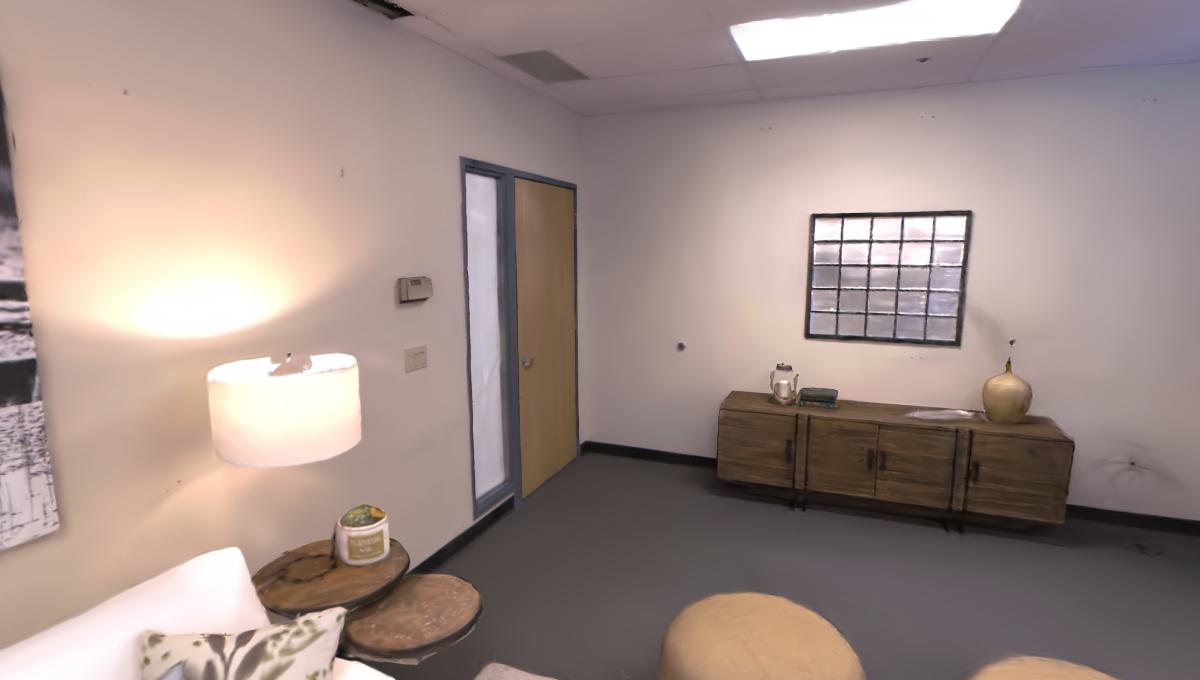}} &
            \drawdottedbox{\reprects}{\includegraphics[width=0.2\linewidth,trim={0px 0px 293px 0px},clip]{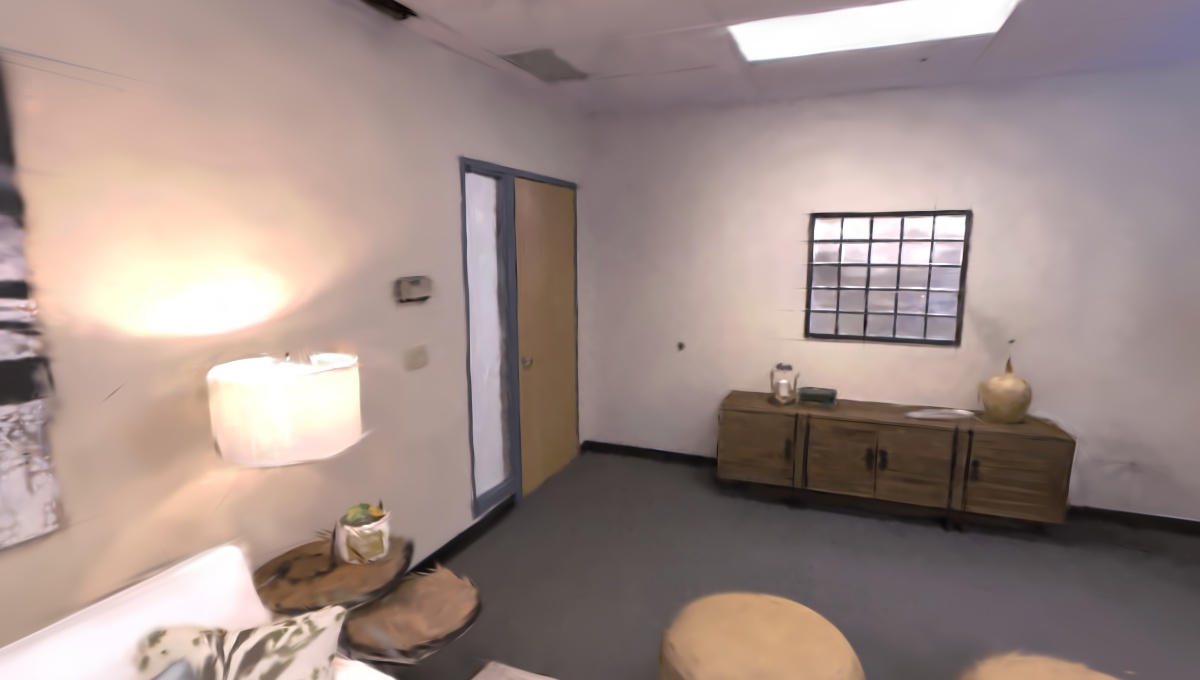}} &
            \includegraphics[width=0.2\linewidth,trim={0px 0px 293px 0px},clip]{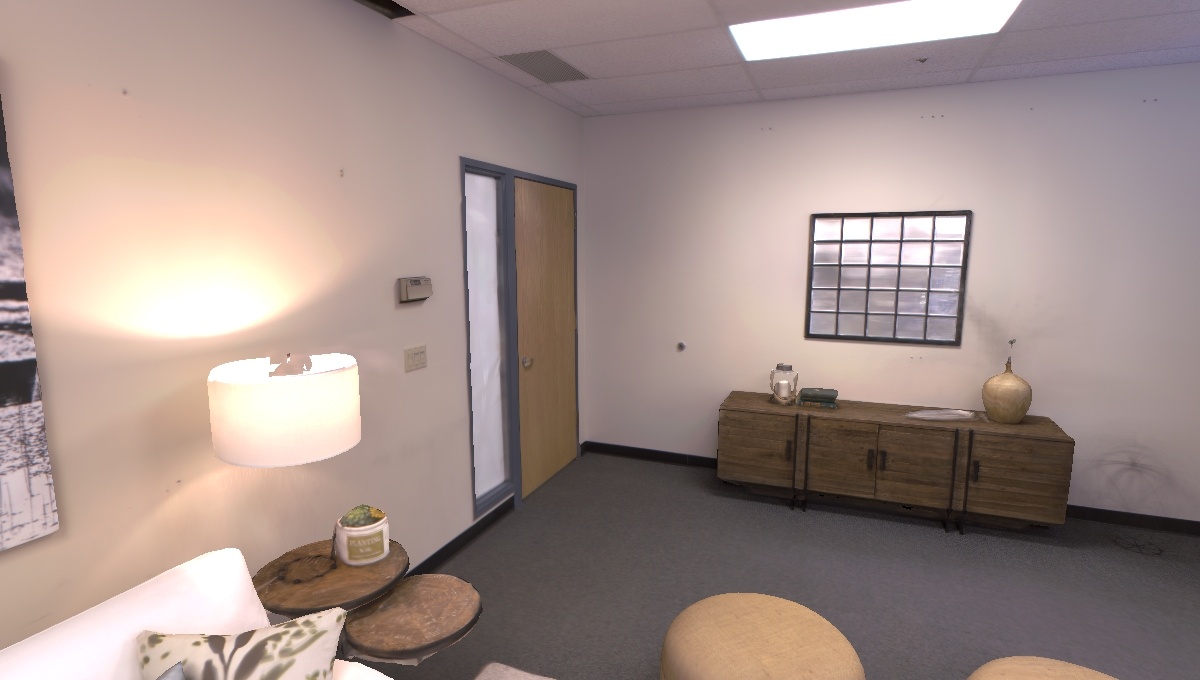} \\
            %
            \drawdottedbox{\tumrects}{\includegraphics[width=0.2\linewidth]{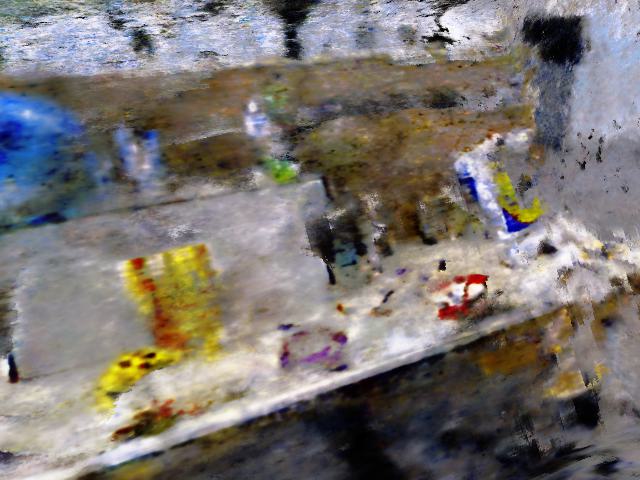}} &
            \drawdottedbox{\tumrects}{\includegraphics[width=0.2\linewidth]{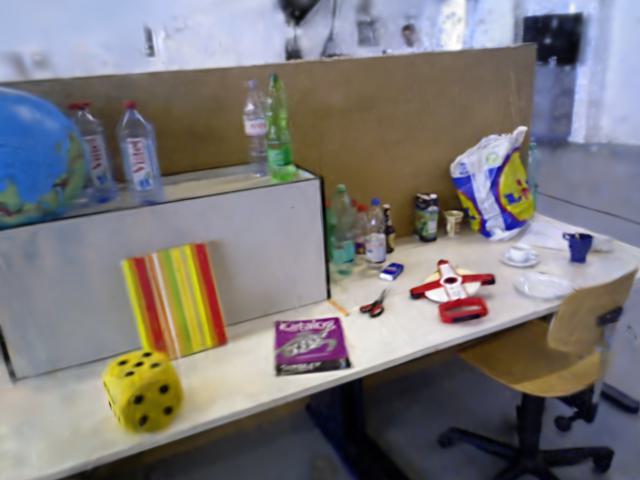}} &
            \drawdottedbox{\tumrects}{\includegraphics[width=0.2\linewidth]{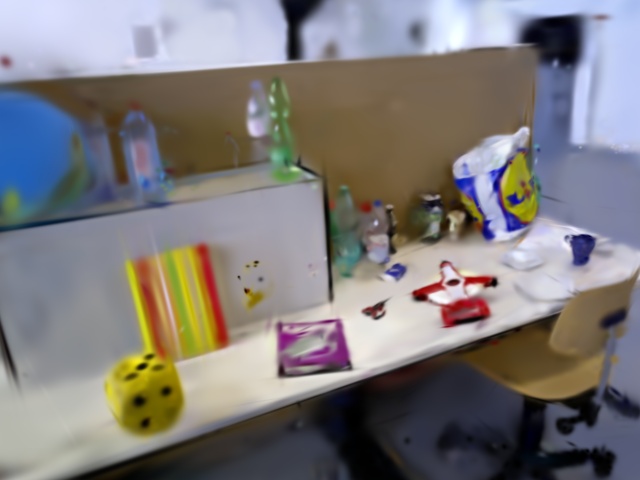}} &
            \drawdottedbox{\tumrects}{\includegraphics[width=0.2\linewidth]{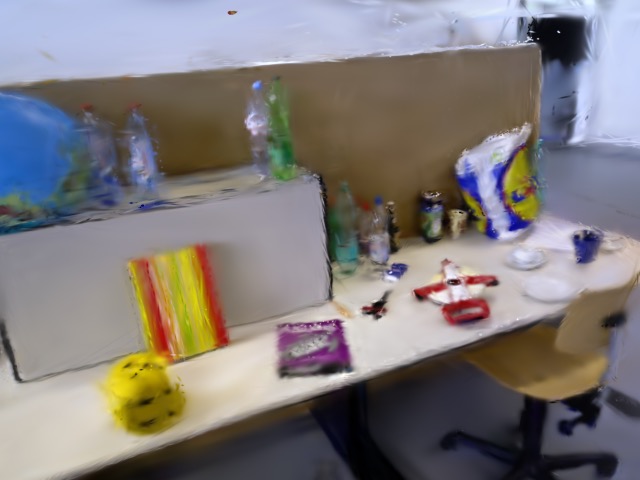}} &
            \drawdottedbox{\tumrects}{\includegraphics[width=0.2\linewidth]{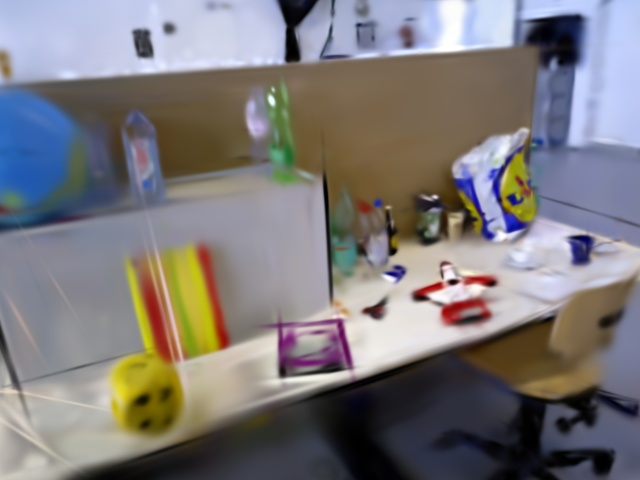}} &
            \includegraphics[width=0.2\linewidth]{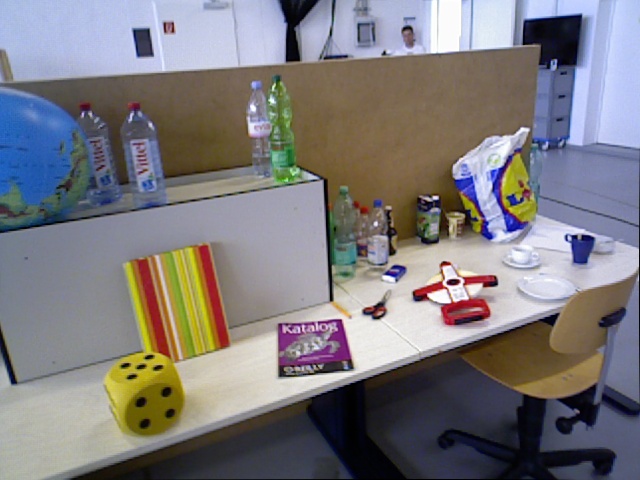} \\
            GO-SLAM & MonoGS & \textbf{Ours (Direct}) & Photo-SLAM & \textbf{Ours (ORB-SLAM}) & Groundtruth
        \end{tabular}
    }
    \caption{GEVO achieves comparable rendering accuracy with other monocular methods on Replica (top) and TUM-RGBD (bottom). 
    \reviewdiff{In particular, GEVO achieves high fidelity by reducing RO, especially for Gaussians representing distant and/or large objects (in green rectangles). Since rendered images degrade in quality slowly over time, optimizating Gaussians using these images in GEVO leads to minor loss of details in some close-up, feature-rich regions (in red rectangles).}
    %
    }
    \label{fig:visualizations}
    \maybevspace{-16pt}
\end{figure*}

\subsubsection{Global Stage}~\label{subsubsec:global-mapping}
%
To resolve remaining inconsistencies of the local map $\widetilde{\mathcal{M}}_{t}$ with prior measurements, we merge Gaussians $\widetilde{\mathcal{M}}_{t}$ into the global map $\mathcal{M}_{t - 1}$. 
During the merging process, we reduce catastrophic forgetting by identifying and updating a small subset of the Gaussians that are temporally inconsistent across past viewpoints by using keyframes \emph{rendered} from these viewpoints.
Since the global map $\mathcal{M}_{t - 1}$ is not perturbed during the local stage, these rendered keyframes from $\mathcal{M}_{t - 1}$ maintain high quality and are sufficient for resolving the remaining inconsistencies. 
%

The global stage consists of the following three sequential steps: insertion, selective optimization, and pruning. 
%

\emph{Insertion and activation}: We insert the local map $\widetilde{\mathcal{M}}_{t}$ and the previous global map $\mathcal{M}_{t - 1}$ to create the pre-optimzied global map: $\mathcal{M}'_{t} \leftarrow \widetilde{\mathcal{M}}_{t} \cup \mathcal{M}_{t - 1}$.
To prevent retrospective occlusion (RO) caused by local map $\widetilde{\mathcal{M}}_{t}$, we lower its opacity to 0.2 prior to the insertion.

\emph{Selective optimization}: We employ an optimization procedure similar to \Cref{eqn:local-mapping} but with two modifications: 1) To prevent Gaussians in the pre-optimized map $\mathcal{M}'_{t}$ from overfitting to the images from the current sliding window buffer $\mathcal{W}_t$, we select and only optimize an active subset $\mathcal{A}_{t} \subseteq \mathcal{M}'_{t}$, and 2) we additionally introduce a photometric consistency loss $E_{pc}$ to further ensure consistency with the prior global map $\mathcal{M}_{t - 1}$.
The overall objective is therefore:
\begin{equation} \label{eqn:global-mapping}
    \argmin_{\mathbf{T}_k, \mathcal{A}_t} E_{pho} + E_{iso} + E_{pc}.
\end{equation}

Specifically, we choose $\mathcal{A}_t = \widetilde{\mathcal{M}}_{t} \cup \mathcal{E}_t$, which contains the newly inserted Gaussians and a subset $\mathcal{E}_t$ that incurs high rendering error in the current window:
\begin{equation}
    \mathcal{E}_t = \left\{ g \in \mathcal{M}_{t - 1}: \max_{k \in \mathcal{W}_t} E_k (g) > \epsilon \right\},
\end{equation}
where the per-Gaussian rendering error $E_k (g)$ as in \cite{bulo2024revising}:
\begin{equation}
    E_k (g) = \sum_{\mathbf{u}} w(g, \mathbf{u}) \abs{R(\mathcal{M}_{t - 1}, \mathbf{T}_k)(\mathbf{u}) - C_k(\mathbf{u})},
\end{equation}
with $\mathbf{u}$ being the pixel coordinate, and $w(g, \mathbf{u})$ is the alpha-blending coefficient of $g$ at pixel $\mathbf{u}$.
Since the Gaussians that are more visible to the camera contribute more to $E_k (g)$, the active set tends to exclude existing Gaussians that are not well-observed by the current window and reduce IRO.

To further ensure that the fidelity of the global map does not degrade over time, the objective in \Cref{eqn:global-mapping} includes a photometric consistency loss $E_{pc}$ evaluated on keyframes at four past camera views outside the sliding window:
\begin{equation}
    E_{pc} = \sum_{l = 1}^4 \norm{R(\mathcal{M}'_t, \mathbf{T}_{k_l}) - \cancel{\nabla} \left( R(\mathcal{M}_{t - 1}, \mathbf{T}_{k_l}) \right)}_1,
\end{equation}
where $R(\cdot)$ is the rendering function, frame index $k_l$ is uniformly sampled from past timesteps $1 \dots t - W$, and $\cancel{\nabla}(\cdot)$ is the stop gradient operator \cite{bulo2024revising}, which avoids back-propagating gradient through $R(\mathcal{M}_{t - 1}, \mathbf{T}_{k_l})$.
%

\emph{Pruning}: After selective optimization, remaining Gaussians that have a) an opacity less than 0.7 or b) an occupancy probability less than 0.9 are likely to cause RO and thus pruned.
The occupancy probability is computed using Gaussian Mixture Regression~\cite{gmmap} on the initialized Gaussians representing free regions from \Cref{subsec:initialization}.

\section{Experiments}~\label{sec:experiments}

\begin{table*}[!t]
    \fixTableTopMargin
    \newcommand{\metriccline}{\cdashline{2-15}}
    \setlength{\aboverulesep}{0pt}
    \renewcommand{\arraystretch}{1.2}
    \setlength{\tabcolsep}{3pt}
    \centering
    \caption{Memory usage and rendering accuracy of GEVO compared with prior works on scenes from the Replica~\cite{straub2019replica} and TUM RGB-D~\cite{sturm12iros} datasets. The best results are highlighted as \colorbox{colorFst}{\bf first} and \colorbox{colorSnd}{second}.}
    \resizebox{\textwidth}{!}{
        \begin{threeparttable}
            \begin{tabular}{cc|rrrrrrrrr|rrrr}
\toprule
Metrics & Methods & \multicolumn{1}{c}{office0} & \multicolumn{1}{c}{office1} & \multicolumn{1}{c}{office2} & \multicolumn{1}{c}{office3} & \multicolumn{1}{c}{office4} & \multicolumn{1}{c}{room0} & \multicolumn{1}{c}{room1} & \multicolumn{1}{c}{room2} & \multicolumn{1}{c|}{Average} & \multicolumn{1}{c}{fr1\_desk} & \multicolumn{1}{c}{fr2\_xyz} & \multicolumn{1}{c}{fr3\_office} & \multicolumn{1}{c}{Average} \\ \hline
\multirow{5}{*}{\shortstack{Overhead Memory\\(MB) $\downarrow$}} & GO-SLAM & {2673.6} & {2794.6} & {3066.2} & {3001.3} & {2575.7} & {3606.0} & {3651.6} & {3796.0} & {3145.6} & {3747.0} & {2870.8} & {4087.5} & {3568.5} \\ \cdashline{2-15} 
 & MonoGS & \marktrd{534.6} & \marktrd{618.6} & {664.8} & {678.4} & \marktrd{383.2} & {697.0} & \marktrd{168.2} & {683.7} & \marktrd{553.6} & \marktrd{106.7} & \marktrd{96.3} & \marktrd{213.3} & \marktrd{138.8} \\
 & \textbf{Ours (Direct)} & \markfst{72.8} & \markfst{71.1} & \markfst{70.5} & \markfst{75.5} & \markfst{67.6} & \markfst{82.3} & \markfst{71.2} & \markfst{71.6} & \markfst{72.8} & \markfst{25.1} & \markfst{24.2} & \markfst{24.3} & \markfst{24.6} \\ \cdashline{2-15} 
 & Photo-SLAM & {606.2} & {1504.7} & \marktrd{631.1} & \marktrd{559.2} & {633.3} & \marktrd{675.1} & {791.6} & \marktrd{650.5} & {756.5} & {135.6} & {336.9} & {629.3} & {367.3} \\
 & \textbf{Ours (ORB-SLAM)} & \marksnd{85.6} & \marksnd{86.1} & \marksnd{100.8} & \marksnd{93.4} & \marksnd{100.3} & \marksnd{92.3} & \marksnd{95.6} & \marksnd{98.5} & \marksnd{94.1} & \marksnd{40.4} & \marksnd{30.7} & \marksnd{43.4} & \marksnd{38.1} \\ \hline
\multirow{5}{*}{\shortstack{Map Memory\\(MB) $\downarrow$}} & GO-SLAM & {48.1} & {48.1} & {48.1} & {48.1} & {48.1} & {48.1} & {48.1} & {48.1} & {48.1} & {48.1} & {48.1} & {48.1} & {48.1} \\ \cdashline{2-15} 
 & MonoGS & \markfst{2.8} & \marksnd{3.1} & \markfst{4.3} & {6.2} & \markfst{2.4} & \marksnd{9.1} & \markfst{2.9} & \marksnd{4.4} & \markfst{4.4} & \marktrd{1.3} & \marktrd{2.4} & \marktrd{2.1} & \marktrd{1.9} \\
 & \textbf{Ours (Direct)} & \marktrd{4.6} & \marktrd{3.4} & {6.6} & \marktrd{5.9} & \marksnd{4.5} & \marktrd{9.1} & \marktrd{5.8} & {5.7} & \marktrd{5.7} & \markfst{0.8} & \markfst{0.4} & \marksnd{1.0} & \marksnd{0.7} \\ \cdashline{2-15} 
 & Photo-SLAM & {4.8} & {10.2} & \marktrd{6.4} & \markfst{4.5} & {5.0} & {9.4} & {9.0} & \marktrd{5.4} & {6.8} & {2.1} & {3.8} & {5.1} & {3.6} \\
 & \textbf{Ours (ORB-SLAM)} & \marksnd{2.9} & \markfst{2.1} & \marksnd{5.9} & \marksnd{4.9} & \marktrd{4.6} & \markfst{7.3} & \marksnd{5.3} & \markfst{4.4} & \marksnd{4.7} & \marksnd{1.0} & \marksnd{0.4} & \markfst{0.7} & \markfst{0.7} \\ \hline
\multirow{5}{*}{PSNR$^1$ (dB) $\uparrow$} & GO-SLAM & \marktrd{28.4} & {28.2} & {19.6} & {22.4} & \marktrd{23.9} & {20.8} & \marktrd{24.8} & {13.2} & {22.7} & {16.0} & {15.7} & {15.4} & {15.7} \\ \cdashline{2-15} 
 & MonoGS & {25.3} & \marktrd{29.1} & {22.0} & {24.3} & {15.2} & {21.9} & {8.2} & {22.4} & {21.1} & \marktrd{17.8} & {14.1} & \markfst{18.1} & {16.7} \\
 & \textbf{Ours (Direct)} & {24.6} & {28.7} & \marktrd{26.5} & \marktrd{25.5} & {15.9} & \marktrd{24.3} & {20.9} & \marktrd{23.9} & \marktrd{23.8} & {17.6} & \marktrd{19.8} & {15.9} & \marktrd{17.8} \\ \cdashline{2-15} 
 & Photo-SLAM & \markfst{35.8} & \markfst{37.3} & \markfst{30.2} & \markfst{30.9} & \markfst{33.2} & \markfst{29.0} & \markfst{31.0} & \markfst{32.3} & \markfst{32.4} & \markfst{20.5} & \markfst{23.0} & \marksnd{18.0} & \markfst{20.5} \\
 & \textbf{Ours (ORB-SLAM)} & \marksnd{33.4} & \marksnd{34.6} & \marksnd{27.7} & \marksnd{28.1} & \marksnd{27.5} & \marksnd{27.0} & \marksnd{28.4} & \marksnd{28.3} & \marksnd{29.4} & \marksnd{18.5} & \marksnd{19.8} & \marktrd{17.1} & \marksnd{18.5} \\ \hline
\multirow{5}{*}{SSIM$^2$ $\uparrow$} & GO-SLAM & {0.76} & {0.76} & {0.63} & {0.68} & {0.76} & {0.54} & {0.72} & {0.44} & {0.66} & {0.46} & {0.45} & {0.44} & {0.45} \\ \cdashline{2-15} 
 & MonoGS & {0.67} & \marktrd{0.86} & {0.81} & {0.82} & {0.40} & {0.71} & {0.16} & {0.80} & {0.65} & \marksnd{0.66} & {0.54} & \markfst{0.67} & {0.62} \\
 & \textbf{Ours (Direct)} & \marktrd{0.81} & {0.84} & \marktrd{0.85} & \marktrd{0.85} & \marktrd{0.81} & \marktrd{0.78} & \marktrd{0.75} & \marktrd{0.85} & \marktrd{0.82} & {0.63} & \marktrd{0.68} & {0.61} & \marktrd{0.64} \\ \cdashline{2-15} 
 & Photo-SLAM & \markfst{0.95} & \markfst{0.95} & \markfst{0.92} & \markfst{0.91} & \markfst{0.93} & \markfst{0.85} & \markfst{0.90} & \markfst{0.93} & \markfst{0.92} & \markfst{0.73} & \markfst{0.78} & \marksnd{0.65} & \markfst{0.72} \\
 & \textbf{Ours (ORB-SLAM)} & \marksnd{0.91} & \marksnd{0.90} & \marksnd{0.88} & \marksnd{0.88} & \marksnd{0.91} & \marksnd{0.82} & \marksnd{0.86} & \marksnd{0.89} & \marksnd{0.88} & \marktrd{0.65} & \marksnd{0.68} & \marktrd{0.64} & \marksnd{0.66} \\ \hline
\multirow{5}{*}{LPIPS$^3$ $\downarrow$} & GO-SLAM & {0.46} & {0.42} & {0.48} & {0.49} & {0.49} & {0.58} & {0.50} & {0.69} & {0.51} & {0.56} & {0.54} & {0.63} & {0.57} \\ \cdashline{2-15} 
 & MonoGS & {0.40} & {0.28} & {0.36} & {0.27} & {0.59} & {0.37} & {0.78} & {0.38} & {0.43} & \marksnd{0.39} & {0.49} & \marksnd{0.42} & \marksnd{0.43} \\
 & \textbf{Ours (Direct)} & \marktrd{0.35} & \marktrd{0.27} & \marktrd{0.27} & \marktrd{0.23} & \marktrd{0.41} & \marktrd{0.26} & \marktrd{0.39} & \marktrd{0.27} & \marktrd{0.31} & {0.45} & \marksnd{0.43} & {0.56} & {0.48} \\ \cdashline{2-15} 
 & Photo-SLAM & \markfst{0.07} & \markfst{0.06} & \markfst{0.10} & \markfst{0.10} & \markfst{0.08} & \markfst{0.12} & \markfst{0.08} & \markfst{0.08} & \markfst{0.09} & \markfst{0.25} & \markfst{0.14} & \markfst{0.29} & \markfst{0.22} \\
 & \textbf{Ours (ORB-SLAM)} & \marksnd{0.20} & \marksnd{0.21} & \marksnd{0.24} & \marksnd{0.20} & \marksnd{0.21} & \marksnd{0.22} & \marksnd{0.22} & \marksnd{0.20} & \marksnd{0.21} & \marktrd{0.43} & \marktrd{0.44} & \marktrd{0.50} & \marktrd{0.46} \\ \hline
\multirow{5}{*}{ATE (cm) $\downarrow$} & GO-SLAM & \markfst{0.3} & \marktrd{0.7} & \markfst{0.4} & \marksnd{0.5} & \marksnd{0.8} & \marktrd{0.5} & \markfst{0.3} & \marksnd{0.3} & \markfst{0.5} & \marksnd{2.3} & \markfst{0.3} & \marksnd{1.8} & \marksnd{1.5} \\ \cdashline{2-15} 
 & MonoGS & {40.7} & {30.5} & {56.3} & {19.2} & {58.3} & {19.9} & {32.9} & {34.7} & {36.6} & \marktrd{2.9} & {20.8} & {10.4} & {11.4} \\
 & \textbf{Ours (Direct)} & {14.0} & {12.9} & {7.8} & {4.2} & {187.4} & {12.3} & {41.2} & {7.3} & {35.9} & {8.0} & {2.6} & {39.1} & {16.6} \\ \cdashline{2-15} 
 & Photo-SLAM & \marktrd{0.6} & \marksnd{0.4} & \marksnd{1.0} & \markfst{0.5} & \markfst{0.7} & \markfst{0.4} & \marksnd{0.7} & \markfst{0.2} & \marksnd{0.6} & \markfst{1.6} & \marksnd{0.4} & \markfst{1.6} & \markfst{1.2} \\
 & \textbf{Ours (ORB-SLAM)} & \marksnd{0.5} & \markfst{0.3} & \marktrd{1.6} & \marktrd{1.0} & \marktrd{0.9} & \marksnd{0.4} & \marktrd{1.6} & \marktrd{1.0} & \marktrd{0.9} & {2.9} & \marktrd{0.8} & \marktrd{2.9} & \marktrd{2.2} \\ \bottomrule
\end{tabular}
        \reviewdiff{
        \begin{tablenotes}
            \item[1-3] Peak Signal-to-Noise Ratio, Structural Similarity Index Measure \cite{nilsson2020understanding}, and Learned Perceptual Image Patch Similarity \cite{zhang2018perceptual}.
        \end{tablenotes}
        }
        \end{threeparttable}
        \label{tab:main-metrics}
    }
    \maybevspace{-16pt}
\end{table*}

We evaluate GEVO against state-of-the-art (SOTA) monocular dense SLAM frameworks in terms of accuracy and efficiency.
To demonstrate the trade-offs among different system configurations, we choose the following frameworks: GO-SLAM\footnote{[Online]. Available: \href{https://github.com/youmi-zym/GO-SLAM}{\color{blue}https://github.com/youmi-zym/GO-SLAM}}~\cite{zhang2023go} (learning-based tracking + neural-based mapping), MonoGS\footnote{[Online]. Available: \href{https://github.com/muskie82/MonoGS}{\color{blue}https://github.com/muskie82/MonoGS}}~\cite{matsuki2024gaussian} (direct tracking + Gaussian-based mapping), and Photo-SLAM\footnote{[Online]. Available: \href{https://github.com/HuajianUP/Photo-SLAM}{\color{blue}https://github.com/HuajianUP/Photo-SLAM}}~\cite{huang2024photo} (feature-based tracking + Gaussian-based mapping).
Compared with prior methods in multiple environments, GEVO reduces the overhead memory by \textbf{8-145$\times$} with up to 10\% computation overhead while maintaining comparable accuracy (\Cref{fig:visualizations}).

This section is organized as follows. After describing implementation details and dataset selection in \Cref{subsec:exp_setup}, we compare the memory usage (\Cref{subsec:memory-usage}), computational efficiency (\Cref{subsec:comp-efficiency}), rendering and localization accuracy (\Cref{subsec:rendering-accuracy}) against SOTA methods. An ablation study for the design of GEVO is presented in \Cref{subsec:ablation}.

\subsection{Experiment Setup}~\label{subsec:exp_setup}

GEVO is implemented in C++ with CUDA acceleration and can be found at \href{https://github.com/mit-lean/gevo}{\color{blue}https://github.com/mit-lean/gevo}.
We benchmarked GEVO and prior methods with an Intel Xeon Gold 6130 and NVIDIA TITAN RTX GPU.
We use a sliding window buffer that stores either 8 (TUM) or 10 (Replica) keyframes.
%
%
%
For prior methods, we use the default settings from the open-source code release for supported datasets or otherwise perform fine-tuning from default settings.

Our method is compatible with various tracking methods.
For fairness, we present results of two variants of GEVO: \emph{Ours (Direct)} employs the photometric tracking from MonoGS whereas \emph{Ours (ORB-SLAM)} uses ORB-SLAM~\cite{campos2021orb} for tracking as in Photo-SLAM.
%
%
For Photo-SLAM and Ours~(ORB-SLAM), we disabled loop closure and downsampled the ORB vocabulary to 1/100 of the original to reduce memory usage without sacrificing accuracy.
%
We disabled spherical harmonics for all GS-based methods.

We benchmarked all frameworks on Replica~\cite{sturm12iros}, a highly detailed synthetic dataset that provides noiseless RGB images for estimating the upper bound performance of the methods, and TUM RGB-D~\cite{straub2019replica}, a real-world dataset for testing the methods under noisy images from a Kinect camera.
Similar to prior works, we select eight sequences from Replica: \texttt{office\_0-4} and \texttt{room\_0-2} and three from TUM RGB-D: \texttt{fr1\_desk}, \texttt{fr2\_xyz} and \texttt{fr3\_office}.

%
%
%

\subsection{Memory Usage}~\label{subsec:memory-usage}

In this section, we compare the memory usage of GEVO against prior frameworks.
The total memory is comprised of: 1) the \emph{map}, which is the size of the resulting NeRF or Gaussians, and 2) the \emph{overhead}, which is the extra memory for storing input and temporal variables during the execution in order to produce the output (the map).
%
%
Note that the overhead memory is highly dependent on the algorithm and its implementation.
To reduce the impact of implementation, we only track variables that are \emph{essential} to the algorithm\footnote{Measured using our own memory profiler that tracks memory allocation of these variables in their C++ constructors and deconstructors.} and avoid ones cached solely for acceleration.

The overhead memory of GEVO and other methods is summarized in \Cref{tab:main-metrics}.
\reviewdiff{Recall that all methods store keyframes in full resolution.}
Overall, GEVO requires the lowest average overhead memory of 83.5~MB on Replica and 31.4~MB on TUM, which are dominated by the keyframes stored in the current window.
In MonoGS and Photo-SLAM, many keyframes from both current and past windows are stored in memory to periodically train Gaussians to reduce catastrophic forgetting.
For these frameworks, the keyframes dominate (96\% to 99\%) the overhead memory.
By only storing images in the current sliding window, GEVO reduces the overhead memory by 114~MB on average on TUM compared with MonoGS and Photo-SLAM.
%
Since the image resolution is $2.66\times$ higher on Replica, GEVO achieves an even higher overhead memory reduction (480~MB on average).

Among the benchmarked methods, GO-SLAM has the highest overhead memory due to the storage of large temporary variables in addition to all keyframe images.
Specifically, with a tracker derived from DROID-SLAM, GO-SLAM computes a 4D correlation volume~\cite{teed2020raft} between each pair of pixels of multiple feature maps to obtain the optical flow.
With a size of at least 1~GB, the 4D correlation volume contributes to 50\% to 60\% of the overhead memory.
Thus, GEVO reduces the overhead memory by up to $53\times$ on Replica and $168\times$ on TUM compared with GO-SLAM.

%
%
%
%

From \Cref{tab:main-metrics}, the map memory of all prior frameworks only consumes a tiny fraction (0.5\% to 2.4\%) of the total memory. Thus, the total memory in each framework is dominated by the overhead memory, which our framework addresses.
Recall that maps in GS-based frameworks (GEVO, MonoGS, and Photo-SLAM) consist of Gaussians whose number scales with the size of the environment.
In contrast, GO-SLAM is a neural-based method whose network size cannot adapt to the size of the environment.
Thus, the map size of GO-SLAM for both datasets is constant at 48~MB, which is consistently 4.7$\times$ to 124$\times$ higher than GS-based frameworks for both datasets.
%
%
%

\reviewdiff{
\subsection{Computational Efficiency}~\label{subsec:comp-efficiency}
GEVO reduces memory overhead at the cost of a negligible increase in computation.
%
%
To measure the compute overhead, we compare GEVO (Direct) and MonoGS in \emph{single-threaded} CPU mode\footnote{\reviewdiff{Aside from our techniques for reducing catastrophic forgetting, GEVO (Direct) deploys a similar computational pipeline and the same GPU rasterizer as MonoGS.}} such that latency is proportional to the compute and is not hidden by multi-threading.

The average latency per image for GEVO is 340~ms and 621~ms per image on TUM and Replica, respectively, which is only 8\% to 10\% higher than MonoGS.
In both works, rasterization dominates up to 70\% of the computation time.
Since GEVO only optimizes Gaussians in the active set (less than 15\% of all Gaussians), a specialized rasterizer can be implemented in the future to greatly reduce overall latency.

}

\subsection{Rendering and Localization Accuracy}~\label{subsec:rendering-accuracy}
We compare the accuracy of GEVO against prior methods by computing the PSNR, \reviewdiff{SSIM~\cite{nilsson2020understanding}, and LPIPS~\cite{zhang2018perceptual}}\footnote{\reviewdiff{PSNR, SSIM, and LPIPS measure the fidelity, texture and overall perceptual differences between the rendered and original image, respectively.}}  metrics of every five non-keyframes in addition to the RMSE of average translation error (ATE) of keyframes.
From \Cref{tab:main-metrics}, GEVO (ORB-SLAM) achieves the second-best rendering and comparable localization accuracy on almost all sequences compared with the best-performing Photo-SLAM.
Our PSNR degrades by 3~dB on Replica and 2~dB on TUM compared with Photo-SLAM
\reviewdiff{due to the usage of rendered images, which are lower fidelity than original ones from camera.}

Since our framework is compatible with many localization methods, we configure our framework using the same direct localization method as MonoGS for its comparison.
Due to a more geometrically verified Gaussian initialization procedure based on traditional multi-view stereo, our framework is less likely to incur a large ATE compared with MonoGS, which randomly initializes Gaussians using priors from the global map.
Due to our consistency-aware optimization, our framework often yields a better rendering accuracy in more than half of the sequences compared with MonoGS without training on up to 300 keyframes stored in memory.
%

\Cref{fig:visualizations} shows sample rendering from the benchmarked sequences.
The rendering of GEVO shows minimal visual difference compared with Photo-SLAM and suffers from fewer artifacts than MonoGS.
%
Despite robust localization, GO-SLAM shows the most noticeable artifacts, especially on TUM, which contains motion blur and lighting changes.

\begin{table}[!t]
    \fixTableTopMargin
	\setlength{\aboverulesep}{0pt}
	\renewcommand{\arraystretch}{1.2}
	\setlength{\tabcolsep}{3pt}
	\centering
	\caption{Impacts of various techniques in GEVO on the PSNR for the TUM RGB-D dataset.}
		\begin{threeparttable}
\setlength{\aboverulesep}{0pt}
\setlength{\belowrulesep}{0pt}
\begin{tabular}{cc|ccccccc}
\toprule
\multicolumn{2}{c|}{Stored References} &  &  &  &  &  &  & \checkmark \\
\multicolumn{2}{c|}{Rendered References} &  & \checkmark &  & \checkmark & \checkmark & \checkmark &  \\
\multicolumn{2}{c|}{Two-Stage Optimization} &  &  & \checkmark & \checkmark & \checkmark & \checkmark &  \\
\multicolumn{2}{c|}{Selective Optimization} &  &  &  &  & \checkmark & \checkmark &  \\
\multicolumn{2}{c|}{Free Space Pruning} &  &  &  &  &  & \checkmark &  \\ \hline
\multirow{2}{*}{PSNR$^*$ (dB)} & Initial & 25.9 & 25.1 & 23.8 & 23.7 & 22.7 & 22.6 & 25.0 \\
 & Final & 15.5 & 17.0 & 16.5 & 18.4 & 18.8 & 19.0 & 19.4 \\ \bottomrule
\end{tabular}
			\begin{tablenotes}
				\item[*] To avoid the impacts of localization, ground truth poses are used.
                \end{tablenotes}
		\end{threeparttable}
		\label{tab:ablation}
        \maybevspace{-16pt}
\end{table}

\subsection{Ablation Studies}~\label{subsec:ablation}

In this section, we show the impact of our proposed techniques on alleviating catastrophic forgetting.
To measure forgetting, we evaluate the rendering PSNR of each keyframe upon its evacuation from the sliding window and after the full sequence is processed, which we label as the initial and final PSNR, respectively.
\Cref{tab:ablation} shows the initial and final PSNR averaged across all keyframes of three sequences in the TUM dataset.
To avoid the impact of localization on map fidelity, we used the ground truth trajectory. 
\reviewdiff{Since the reduced resolution of the cost volume during Gaussian initialization has negligible effect on PSNR (\ie, less than 0.2 dB), we perform ablation study on remaining techniques.}

Without retraining with any images outside the current sliding window (leftmost column of \Cref{tab:ablation}), the rendering quality of keyframes degrades significantly over time due to catastrophic forgetting (see \Cref{fig:forgetting-visualization-last}).
Simply replacing stored images with rendered ones (column 2) alone noticeably increases the final PSNR.
However, since rendered images degrade in fidelity over time, using them alone is not sufficient without additional techniques (columns 3-6).
All techniques combined recover the final average PSNR to 19 dB (see \Cref{fig:forgetting-visualization-ours}), which is only 0.4~dB lower than storing and training with original images (rightmost column).
Although each additional technique gradually lowers the initial PSNR due to stronger regularization, they tend to impede forgetting during optimization.
Thus, all our techniques combined recover most of the loss in fidelity due to not storing and retraining using the original past images.

\section{Conclusion}
In this letter, we presented GEVO, a memory-efficient GS-based monocular SLAM that avoids catastrophic forgetting due to incomplete sensor obscuration (IRO) and retrospective occlusion (RO) without storing past images.
%
%
By using rendered images to guide the optimization and introducing occupancy-preserving initialization and consistency-aware optimization to retain their fidelity, map consistency is maintained.
%
Experiments on the TUM and Replica datasets show that while maintaining comparable rendering accuracy with SOTA methods, GEVO reduces the memory overhead to 58~MBs, which is up to 94$\times$ lower than prior methods.
Thus, GEVO has made a significant step towards deploying GS-based SLAM on energy-constrained devices. 


{
\bibliographystyle{IEEEtran}
\bibliography{references}

\begin{thebibliography}{10}
\providecommand{\url}[1]{#1}
\csname url@rmstyle\endcsname
\providecommand{\newblock}{\relax}
\providecommand{\bibinfo}[2]{#2}
\providecommand\BIBentrySTDinterwordspacing{\spaceskip=0pt\relax}
\providecommand\BIBentryALTinterwordstretchfactor{4}
\providecommand\BIBentryALTinterwordspacing{\spaceskip=\fontdimen2\font plus
\BIBentryALTinterwordstretchfactor\fontdimen3\font minus \fontdimen4\font\relax}
\providecommand\BIBforeignlanguage[2]{{%
\expandafter\ifx\csname l@#1\endcsname\relax
\typeout{** WARNING: IEEEtran.bst: No hyphenation pattern has been}%
\typeout{** loaded for the language `#1'. Using the pattern for}%
\typeout{** the default language instead.}%
\else
\language=\csname l@#1\endcsname
\fi
#2}}

\bibitem{markhoro}
M.~Horowitz, ``1.1 computing's energy problem (and what we can do about it),'' in \emph{2014 IEEE International Solid-State Circuits Conference Digest of Technical Papers (ISSCC)}, 2014, pp. 10--14.

\bibitem{matsuki2024gaussian}
H.~Matsuki, R.~Murai, P.~H. Kelly, and A.~J. Davison, ``Gaussian splatting slam,'' in \emph{Proceedings of the IEEE/CVF Conference on Computer Vision and Pattern Recognition}, 2024, pp. 18\,039--18\,048.

\bibitem{matsuki2023gaussian}
H.~Matsuki, R.~Murai, P.~H.~J. Kelly, and A.~J. Davison, ``{G}aussian {S}platting {SLAM},'' in \emph{Proceedings of the IEEE/CVF Conference on Computer Vision and Pattern Recognition}, 2024.

\bibitem{campos2021orb}
C.~Campos, R.~Elvira, J.~J.~G. Rodr{\'\i}guez, J.~M. Montiel, and J.~D. Tard{\'o}s, ``Orb-slam3: An accurate open-source library for visual, visual--inertial, and multimap slam,'' \emph{IEEE Transactions on Robotics}, vol.~37, no.~6, pp. 1874--1890, 2021.

\bibitem{forster2016manifold}
C.~Forster, L.~Carlone, F.~Dellaert, and D.~Scaramuzza, ``On-manifold preintegration for real-time visual--inertial odometry,'' \emph{IEEE Transactions on Robotics}, vol.~33, no.~1, pp. 1--21, 2016.

\bibitem{davison2007monoslam}
A.~J. Davison, I.~D. Reid, N.~D. Molton, and O.~Stasse, ``Monoslam: Real-time single camera slam,'' \emph{IEEE transactions on pattern analysis and machine intelligence}, vol.~29, no.~6, pp. 1052--1067, 2007.

\bibitem{Mur-Artal-RSS-15}
R.~Mur-Artal and J.~Tardos, ``Probabilistic semi-dense mapping from highly accurate feature-based monocular slam,'' in \emph{Proceedings of Robotics: Science and Systems}, Rome, Italy, July 2015.

\bibitem{rosinol2023nerf}
A.~Rosinol, J.~J. Leonard, and L.~Carlone, ``Nerf-slam: Real-time dense monocular slam with neural radiance fields,'' in \emph{2023 IEEE/RSJ International Conference on Intelligent Robots and Systems (IROS)}.\hskip 1em plus 0.5em minus 0.4em\relax IEEE, 2023, pp. 3437--3444.

\bibitem{matsuki2023imode}
H.~Matsuki, E.~Sucar, T.~Laidow, K.~Wada, R.~Scona, and A.~J. Davison, ``imode: Real-time incremental monocular dense mapping using neural field,'' in \emph{2023 IEEE International Conference on Robotics and Automation (ICRA)}.\hskip 1em plus 0.5em minus 0.4em\relax IEEE, 2023, pp. 4171--4177.

\bibitem{zhu2024nicer}
Z.~Zhu, S.~Peng, V.~Larsson, Z.~Cui, M.~R. Oswald, A.~Geiger, and M.~Pollefeys, ``Nicer-slam: Neural implicit scene encoding for rgb slam,'' in \emph{2024 International Conference on 3D Vision (3DV)}.\hskip 1em plus 0.5em minus 0.4em\relax IEEE, 2024, pp. 42--52.

\bibitem{huang2023photo}
H.~Huang, L.~Li, C.~Hui, and S.-K. Yeung, ``Photo-slam: Real-time simultaneous localization and photorealistic mapping for monocular, stereo, and rgb-d cameras,'' in \emph{Proceedings of the IEEE/CVF Conference on Computer Vision and Pattern Recognition}, 2024.

\bibitem{kerbl20233d}
B.~Kerbl, G.~Kopanas, T.~Leimk{\"u}hler, and G.~Drettakis, ``3d gaussian splatting for real-time radiance field rendering,'' \emph{ACM Transactions on Graphics (ToG)}, vol.~42, no.~4, pp. 1--14, 2023.

\bibitem{lowe2004sift}
G.~Lowe, ``Sift-the scale invariant feature transform,'' \emph{Int. J}, vol.~2, no. 91-110, p.~2, 2004.

\bibitem{rublee2011orb}
E.~Rublee, V.~Rabaud, K.~Konolige, and G.~Bradski, ``Orb: An efficient alternative to sift or surf,'' in \emph{2011 International conference on computer vision}.\hskip 1em plus 0.5em minus 0.4em\relax Ieee, 2011, pp. 2564--2571.

\bibitem{engel2014lsd}
J.~Engel, T.~Sch{\"o}ps, and D.~Cremers, ``Lsd-slam: Large-scale direct monocular slam,'' in \emph{European conference on computer vision}.\hskip 1em plus 0.5em minus 0.4em\relax Springer, 2014, pp. 834--849.

\bibitem{dtam}
R.~A. Newcombe, S.~J. Lovegrove, and A.~J. Davison, ``Dtam: Dense tracking and mapping in real-time,'' in \emph{2011 International Conference on Computer Vision}, 2011, pp. 2320--2327.

\bibitem{ptam}
G.~Klein and D.~Murray, ``Parallel tracking and mapping for small ar workspaces,'' in \emph{2007 6th IEEE and ACM international symposium on mixed and augmented reality}.\hskip 1em plus 0.5em minus 0.4em\relax IEEE, 2007, pp. 225--234.

\bibitem{dso}
J.~Engel, V.~Koltun, and D.~Cremers, ``Direct sparse odometry,'' \emph{IEEE transactions on pattern analysis and machine intelligence}, vol.~40, no.~3, pp. 611--625, 2017.

\bibitem{yang2019monocular}
S.~Yang and S.~Scherer, ``Monocular object and plane slam in structured environments,'' \emph{IEEE Robotics and Automation Letters}, vol.~4, no.~4, pp. 3145--3152, 2019.

\bibitem{nicholson2018quadricslam}
L.~Nicholson, M.~Milford, and N.~S{\"u}nderhauf, ``Quadricslam: Dual quadrics from object detections as landmarks in object-oriented slam,'' \emph{IEEE Robotics and Automation Letters}, vol.~4, no.~1, pp. 1--8, 2018.

\bibitem{rosinol2020kimera}
A.~Rosinol, M.~Abate, Y.~Chang, and L.~Carlone, ``Kimera: an open-source library for real-time metric-semantic localization and mapping,'' in \emph{2020 IEEE International Conference on Robotics and Automation (ICRA)}.\hskip 1em plus 0.5em minus 0.4em\relax IEEE, 2020, pp. 1689--1696.

\bibitem{zhang2023go}
Y.~Zhang, F.~Tosi, S.~Mattoccia, and M.~Poggi, ``Go-slam: Global optimization for consistent 3d instant reconstruction,'' in \emph{Proceedings of the IEEE/CVF International Conference on Computer Vision}, 2023, pp. 3727--3737.

\bibitem{huang2024photo}
H.~Huang, L.~Li, H.~Cheng, and S.-K. Yeung, ``Photo-slam: Real-time simultaneous localization and photorealistic mapping for monocular stereo and rgb-d cameras,'' in \emph{Proceedings of the IEEE/CVF Conference on Computer Vision and Pattern Recognition}, 2024, pp. 21\,584--21\,593.

\bibitem{keetha2024splatam}
N.~Keetha, J.~Karhade, K.~M. Jatavallabhula, G.~Yang, S.~Scherer, D.~Ramanan, and J.~Luiten, ``Splatam: Splat, track \& map 3d gaussians for dense rgb-d slam,'' in \emph{Proceedings of the IEEE/CVF Conference on Computer Vision and Pattern Recognition}, 2024.

\bibitem{gmmap}
P.~Z.~X. Li, S.~Karaman, and V.~Sze, ``Gmmap: Memory-efficient continuous occupancy map using gaussian mixture model,'' \emph{IEEE Transactions on Robotics}, vol.~40, pp. 1339--1355, 2024.

\bibitem{wang2018quadtree}
K.~Wang, W.~Ding, and S.~Shen, ``Quadtree-accelerated real-time monocular dense mapping,'' in \emph{2018 IEEE/RSJ International Conference on Intelligent Robots and Systems (IROS)}.\hskip 1em plus 0.5em minus 0.4em\relax IEEE, 2018, pp. 1--9.

\bibitem{schoenberger2016sfm}
J.~L. Sch\"{o}nberger and J.-M. Frahm, ``Structure-from-motion revisited,'' in \emph{Conference on Computer Vision and Pattern Recognition (CVPR)}, 2016.

\bibitem{schoenberger2016mvs}
J.~L. Sch\"{o}nberger, E.~Zheng, M.~Pollefeys, and J.-M. Frahm, ``Pixelwise view selection for unstructured multi-view stereo,'' in \emph{European Conference on Computer Vision (ECCV)}, 2016.

\bibitem{felzenszwalb2006efficient}
P.~F. Felzenszwalb and D.~P. Huttenlocher, ``Efficient belief propagation for early vision,'' \emph{International journal of computer vision}, vol.~70, pp. 41--54, 2006.

\bibitem{min2014fast}
D.~Min, S.~Choi, J.~Lu, B.~Ham, K.~Sohn, and M.~N. Do, ``Fast global image smoothing based on weighted least squares,'' \emph{IEEE Transactions on Image Processing}, vol.~23, no.~12, pp. 5638--5653, 2014.

\bibitem{eckart2016accelerated}
B.~Eckart, K.~Kim, A.~Troccoli, A.~Kelly, and J.~Kautz, ``Accelerated generative models for 3d point cloud data,'' in \emph{Proceedings of the IEEE conference on computer vision and pattern recognition}, 2016, pp. 5497--5505.

\bibitem{o2018variable}
C.~O’Meadhra, W.~Tabib, and N.~Michael, ``Variable resolution occupancy mapping using gaussian mixture models,'' \emph{IEEE Robotics and Automation Letters}, vol.~4, no.~2, pp. 2015--2022, 2018.

\bibitem{dhawale2020efficient}
A.~Dhawale and N.~Michael, ``Efficient parametric multi-fidelity surface mapping,'' in \emph{Robotics: Science and Systems (RSS)}, vol.~2, no.~3, 2020, p.~5.

\bibitem{goel2023probabilistic}
K.~Goel, N.~Michael, and W.~Tabib, ``Probabilistic point cloud modeling via self-organizing gaussian mixture models,'' \emph{IEEE Robotics and Automation Letters}, vol.~8, no.~5, pp. 2526--2533, 2023.

\bibitem{bulo2024revising}
S.~R. Bul{\`o}, L.~Porzi, and P.~Kontschieder, ``Revising densification in gaussian splatting,'' \emph{arXiv preprint arXiv:2404.06109}, 2024.

\bibitem{straub2019replica}
J.~Straub, T.~Whelan, L.~Ma, Y.~Chen, E.~Wijmans, S.~Green, J.~J. Engel, R.~Mur-Artal, C.~Ren, S.~Verma, \emph{et~al.}, ``The replica dataset: A digital replica of indoor spaces,'' \emph{arXiv preprint arXiv:1906.05797}, 2019.

\bibitem{sturm12iros}
J.~Sturm, N.~Engelhard, F.~Endres, W.~Burgard, and D.~Cremers, ``A benchmark for the evaluation of rgb-d slam systems,'' in \emph{Proc. of the International Conference on Intelligent Robot Systems (IROS)}, Oct. 2012.

\bibitem{nilsson2020understanding}
J.~Nilsson and T.~Akenine-M{\"o}ller, ``Understanding ssim,'' \emph{arXiv preprint arXiv:2006.13846}, 2020.

\bibitem{zhang2018perceptual}
R.~Zhang, P.~Isola, A.~A. Efros, E.~Shechtman, and O.~Wang, ``The unreasonable effectiveness of deep features as a perceptual metric,'' in \emph{CVPR}, 2018.

\bibitem{teed2020raft}
Z.~Teed and J.~Deng, ``Raft: Recurrent all-pairs field transforms for optical flow,'' in \emph{Computer Vision--ECCV 2020: 16th European Conference, Glasgow, UK, August 23--28, 2020, Proceedings, Part II 16}.\hskip 1em plus 0.5em minus 0.4em\relax Springer, 2020, pp. 402--419.

\end{thebibliography}
}

\end{document}